# Aggregate effects of advertising decisions: a complex systems look at search engine advertising via an experimental study

Yanwu Yang[1], Xin Li[2], Bernard J. Jansen[3], Daniel Zeng[4]

[1]School of Management, Huazhong University of Science and Technology, Wuhan 430074, China

[2]Department of Information Systems, City University of Hong Kong, Kowloon Tong, Hong Kong

[3]College of Information Sciences and Technology, Pennsylvania State University, PA 16802, United States

[4]Department of Management Information Systems, University of Arizona, Tucson, AZ 85721, United States

**Abstract**

**Purpose**: We model group advertising decisions, which are the collective decisions of every single advertiser within the set of advertisers who are competing in the same auction or vertical industry, and examine resulting market outcomes, via a proposed simulation framework named EXP-SEA (Experimental Platform for Search Engine Advertising) supporting experimental studies of collective behaviors in the context of search engine advertising.

**Design**: We implement the EXP-SEA to validate the proposed simulation framework, also conduct three experimental studies on the aggregate impact of electronic word-of-mouth, the competition level, and strategic bidding behaviors. EXP-SEA supports heterogeneous participants, various auction mechanisms, and also ranking and pricing algorithms.

**Findings**: Findings from our three experiments show that (a) both the market profit and advertising indexes such as number of impressions and number of clicks are larger when the eWOM effect presents, meaning social media certainly has some effect on search engine advertising outcomes, (b) the competition level has a monotonic increasing effect on the market performance, thus search engines have an incentive to encourage both the eWOM among search users and competition among advertisers, and (c) given the market-level effect of the percentage of advertisers employing a dynamic greedy bidding strategy, there is a cut-off point for strategic bidding behaviors.

**Originality**: This is one of the first research works to explore collective group decisions and resulting phenomena in the complex context of search engine advertising via developing and validating a simulation framework that supports assessments of various advertising strategies and estimations of the impact of mechanisms on the search market.

## 1. Introduction

Search engines are one of the major venues for people to find information online and for advertisers to distribute information concerning their products. One representative advertising model for search engines is search engine advertising (a.k.a. sponsored search or search advertising), where search engines deliver advertisements to searchers according to their submitted queries. Search engine users often have specific objectives (Orso et al. 2017), which make searchers a perfect population for targeted keyword marketing (Yang et al. 2017; Yan et al. 2017; Ramaboa & Fish 2018). According to the IAB's Internet Advertising Revenue Report (IAB 2016), search engine advertising accounts for 48% of the $72.5 billion in annual Internet advertising revenue. It is apparent that search engine advertising is a popular advertising format and the most important revenue source for search engines.

To more fully understand search engine advertising, many research efforts have been invested in model search user behaviors, including queries, clicks, and browsing (Jansen, Booth & Spink 2008; Yao & Mela 2011). However, another research perspective concerns the advertisers, including the advertising strategies of dealing with bid price determination (Zhou, Chakrabarty & Lukose 2008; Zhang & Feng 2011), budget allocation (Yang et al. 2014; Yang et al. 2015), and keyword selection (Abhishek & Hosanagar 2007; Bartz, Murthi & Sebastian 2006). Existing research typically focuses on an individual advertiser, with minimal attention to group advertising behaviors, group advertising decisions, and resulting aggregate effects. A given change of auction mechanisms will affect each individual advertiser and the market differently; moreover, the outcome of the group adoption of a specific advertising strategy is definitely distinct from that of a single advertiser. It is this group advertising decision that is the focus of our research to fill this gap in the search engine advertising literature. We define group advertising decision as the collective decisions of every single advertiser within the set of advertisers that are competing in the same auction or vertical industry.

There are several challenges to studying group advertising decisions. First, in most instances, researchers have no access to advertising datasets involving a large number of advertisers and have no opportunities to conduct field experiments concerning group advertising behaviors. Second, due to the nature of ascending auctions (Edelman, Ostrovsky & Schwarz 2007; Varian 2007) and the dynamic environment of the search engine advertising market, managing both advertising campaigns (for individual advertisers) and auction platforms (for search engines) is quite complicated. This situation becomes even more complex given that one must account for the decisions of the other advertisers in the market. Third, in search engine advertising, interactions among search users, advertisers, and various advertised objects (e.g., campaigns, adgroups, adcopies, and keywords) are interwoven. In effect, due to complicated relationships among involved advertising participants and components, the search engine advertising market has evolved into a complex system where the emerging market phenomena are difficult to analytically represent and predict a priori. The objective of this research is to study the outcome of advertising decisions at the aggregate level (i.e., group advertising decisions) in search engine advertising. To the best of our knowledge, this is the first research in this direction.

This research proposes a simulation framework to support experimental studies on the effects of group advertising decisions on the market outcome in search engine advertising. This framework is based on a multi-agent modeling technology, i.e., the artificial society, where search users and advertisers are represented with various agents. Through their interactions, the artificial society can be grown up from the bottom for experimental studies on search engine advertising. The artificial society approach has potentials in studying complex systems, especially in understanding group phenomena that emerges through individuals' behaviors (Lorini, Perrussel & Mühlenbernd 2016). This framework supports heterogeneous users, various auction mechanisms, and ranking and pricing algorithms. It allows advertisers to derive assessments of various advertising strategies and supports search providers to estimate the impact of mechanisms on the search market by inspecting the collective phenomena that emerge from large-scale simulations. Moreover, we also implement this framework in a prototype system named EXP-SEA (Experimental Platform for Search Engine Advertising).

Based on the EXP-SEA, we conducted three experimental studies. The first experiment examines the effect of the electronic word-of-mouth (eWOM) on the search market, and experimental results show that both market profit and other advertising indexes (i.e., number of impressions, clicks, and actions, the click-through-rate, the conversion rate, and the cost-per-click) become larger when the eWOM effect presents. This finding is in line with the tendency of integrating search engine advertising and social media advertising (Lin et al. 2018) as the eWOM effect (Xun & Guo 2017) in the latter complements the former (Jansen et al. 2009). The second experiment explores the effect of the competition level on the search market, and experimental results show that the competition level has a monotonic increasing effect on the market performance. Therefore, as market makers, search engines have a strong motivation to enhance both the eWOM among search users (Mukherjee & Jansen 2017) and the competition among advertisers[1]. The third experiment studies the market-level effect of strategic bidding behaviors (e.g., employing a dynamic bidding strategy), and experimental results show that it has a non-monotonic effect on the market performance. This finding indicates that there is a certain cut-off point for advertisers in a market segment who adopt certain bidding strategies. Search engine advertising providers need to be cautious about this cut-off point for better market performance. Thus, our simulation framework can help search engines and advertisers understand the effect of market mechanisms and can be used to compare the performance of different advertising strategies. Such a simulation framework will be potentially valuable for both industrial and academic development in this field.

This research makes a substantive contribution to the literature on search engine advertising. First, from the methodological perspective, we provide an artificial society-based framework that can support experimental studies of collective behaviors with different settings of the underlying processes and strategies in search engine advertising. Second, we conduct three experimental studies (i.e. eWOM, competition level and strategic bidding behaviors), and analyze interesting phenomena of collective activities that illustrate the usefulness of our framework in search engine advertising. Additionally, the findings generated from our experiments can provide useful insights both search engines and advertisers.

The remainder of this paper is organized as follows. Section 2 gives a brief survey of related work, focusing on modeling, simulation, and decision support in search engine advertising; Section 3 presents a simulation framework based on an artificial society of search users and advertisers; in Section 4, we implement an experimental platform based on our simulation framework and report some experimental results; and Section 5 concludes this work.

---

[1] We didn't consider the possible negative response of advertisers to the competition in this research, which is worthwhile to explore in the future work.

## 2. Related Work

### 2.1 Search engine advertising

In search engine advertising, there are a limited number of advertising slots for advertisers to display their advertisements. These advertisements are displayed along with organic search results on a search engine results pages (SERPs) when a search user's query contains certain keywords. The allocation of advertisers to these slots is typically conducted by ascending auctions where the Generalized Second Price (GSP) is widely used by major search engines. We first present a brief background of the current state of search engine advertising research.

Most current research efforts in the field of search engine advertising focus on three categories: search auction mechanism design and equilibrium analysisen, (Chen, Feng & Whinston 2010; Liu, Chen & Whinston 2010), empirical studies on different factors (Ghose & Yang 2009; Yang & Ghose 2010), and strategic advertising behaviors and decisions (Feldman et al. 2007). The first concerns mechanism design and equilibrium analysis to find appropriate incentive-compatible solutions (Varian 2007) either for the maximization of social welfare or -other advertising metrics (Jansen & Clarke 2017). The second category aims to empirically identify relationships among various advertising factors and their effects on advertising outcomes (Li et al. 2016). The third focuses on analyzing advertising behaviors (Agarwal & Mukhopadhyay 2016; Lu, Chau & Chau 2017; Abhishek & Hosanagar 2007) and on developing optimal advertising strategies to facilitate campaign manipulations for advertisers, including bid price determination (Zhou, Chakrabarty & Lukose 2008; Zhang & Feng 2011), budget allocation (Yang et al. 2015; Yang et al. 2014), branding (Jun & Park 2017), and keyword selection (Abhishek & Hosanagar 2007; Bartz, Murthi & Sebastian 2006).

However, from the perspective of advertisers, given the high volume and diversity of search queries, it is impossible to manually determine the optimal strategy in real-time. Therefore, advertisers need dynamic advertising strategies adjusting the real-time advertising performance in a given promotion period, rather than static strategies (Katona & Sarvary 2010; Yang et al. 2012; Yang et al. 2013; Yao & Mela 2011). Although previous studies have developed many advertising strategies, advertisers need more information to quickly determine when to choose which strategy in response to other market participants' moves.

From the perspective of search providers, it is necessary to make decisions on auction setups within the structure of search demand and supply. In addition, the proliferation of click fraud (Jansen 2007) and vindictive bidding strategies (Zhou & Lukose 2007) place demands on major search engines to frequently update and adjust their auction mechanisms in order to maintain their search auction platforms, and to keep exploring novel auction methodologies in order to guarantee their leading positions in the search engine advertising industry.

### 2.2 Simulation and Modeling in Search engine advertising

In order to enable analytical analysis of Nash equilibrium, typically search engine advertising auction is modeled as a one-shot game with complete information and focusing on a single keyword auction (Edelman, Ostrovsky & Schwarz 2007; Varian 2007). Bidding strategies derived from such stylized settings with strictly simplifying assumptions can only be evaluated in terms of theoretical properties (Berg et al. 2010), while bidding behaviors observed in practice reveals strategic uncertainty, which accounts for the departure from the VCG outcome (Che, Choi & Kim 2017).

Consequently, there is quite a long way before delivering such advertising strategies to real applications, leaving room for the development of search engine advertising simulations (Acharya et al. 2007).

Indeed, given its complexity from both advertisers and searchers, the search engine advertising area is a rapidly developing area with a significant need for modeling and simulation (Gupta, Saha & Sarkar 2016). The majority of the prior work on search engine advertising simulation has been conceptual work on the effect of pricing knowledge. For example, Kitts and LeBlanc (2004) developed a bid management framework to query market quotes, store advertising data, and submit bids. Consequently, they constructed an auction simulator to validate their framework and trading agents, claimed that trading agents could estimate unknown factors (such as click volume and position), explore auctions with different bids to infer competitor price points, then seek out the gap and position itself. The growing prosperity of search engine advertising is largely driven by the influx of millions of advertisers, which unavoidably leads to intense advertising competition (Yang et al. 2016). Using game-theoretic analysis framework to provide experimental contexts, Jordan et al. (2007) provided a framework where different strategies are ranked, then observed and analyzed comparing with the tournament ranking. Through post-tournament simulation, an empirical game is constructed to derive equilibriums. Similarly, Jordan et al. (2010) employed a game-theoretic method of equilibrium analysis to identify key strategic behaviors of the competition's top agents. Jordan et al. (2007) work is important as it begins to show the interplay among individual advertisers. In a sequent work, Pardoe et al. (2010) designed individual agents to make predictions on various factors of the game and to bid against each other over the course of simulated auctions, providing insights on the reaction of individual advertisers to changes in the environment. In a similar work, Abrams et al. (2007) also proposed a simulation framework for agent bidding in search engine advertising.

From the perspective of the search engine advertising platform, Lahaie et al. (2007) presented an extensive overview of basic mechanisms and of the search engine advertising process. Showing the implication of simulation research in this area, Acharya et al. (2007) presented a simulation system to evaluate alternative designs and features in search engine advertising markets. The stratified sampling and micro-market sampling strategies generated a small-scale representation of a complete market for effective evaluations, demonstrating that simulation can be effective for market insights.

Despite the advances of these prior works, there have been limited efforts on simulation and modeling to mimic the dynamic environments of search engine advertising and to enable evaluating various mechanisms and strategies at different levels of abstractions (e.g., individual advertiser, the market). To the best of our knowledge, neither of these prior works investigated aggregate aspects of search engine advertising, i.e., the assessment of market-level performance by scalable simulations. This research aims to fill this gap in the literature.

## 3. The Simulation Framework

This work develops an artificial society-based simulation framework to capture the dynamic complexity of search engine advertising, with a foundation in general systems theory (Bertalanffy 1968). Artificial society is an agent-based computational model for social simulation and analysis, consisting of a population of intelligent agents, a set of rules governing agents' interactions, and an environment (Epstein & Axtell 1996). It facilitates relationships and social actions among intelligent agents in order to understand and to simulate adaptive behaviors, properties, emergent complex social structures, and processes from simple interactions of imitation and induction of simple behaviors and rules in individual agents (Sawyer 2003). Thus, it is an ideal methodology for search engine advertising simulation.

### 3.1 System Architecture

Figure 1 shows the architecture of our simulation framework, with three basic ingredients: (a) agents of search users and advertisers, (b) a set of rules for behaviors of agents, and (c) the underlying environment. The upper part is the search-advertising environment containing advertisement retrieval, auction mechanisms, matching rules, ranking, and pricing components.

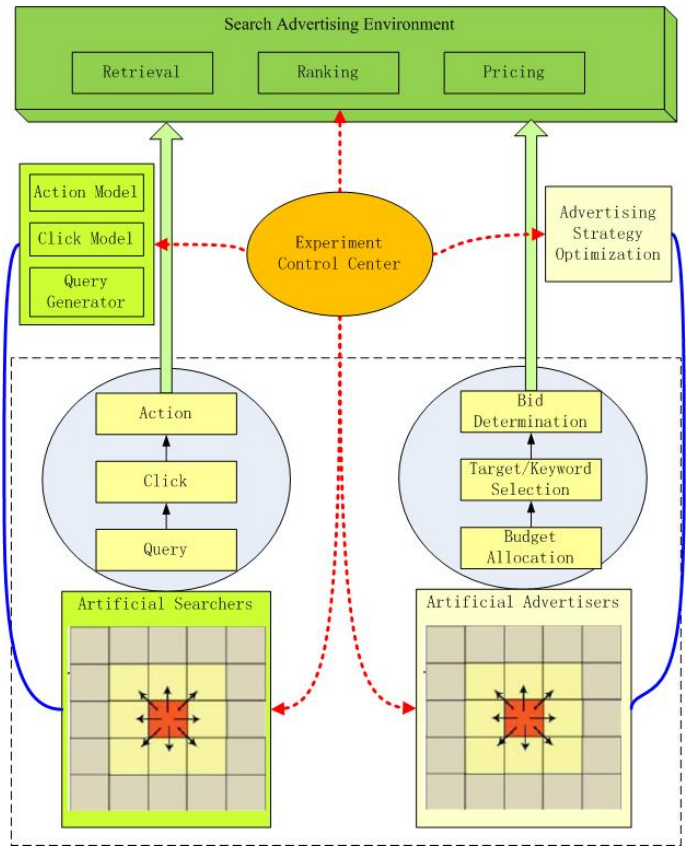

Figure 1. A simulation framework for search engine advertising

The bottom part represents the artificial societies of search users and advertisers. In this framework, an artificial society model is used to describe the various types of participants, namely, search users and advertisers. Search users are specified with a set of behavior models (e.g., query, click, and action), and search advertisers are specified with a set of advertising decisions, such as bid price determination, target/keyword selection, and budget allocation (i.e., the blue solid line). These agents interact with each other following these predefined rules in the underlying environment of search engine advertising. In order to represent the interactions between search users (or between advertisers), we employ the cellular automata to instantiate the artificial society because the cellular automata technique is simple yet powerful for describing complex phenomena and patterns (Wolfram 1984). Without loss of generality, the artificial search users and advertisers are represented in the context of the two-dimensional cellular automata (i.e., these agents interact with each other on a grid as illustrated the bottom of the framework). Although search users and advertisers are both implemented with the grid cell automata, they have different implications. The search users on the left are proactive and initiate various actions, and the advertisers on the right are passive and respond to search users' actions with predefined rules/strategies. Moreover, for a search user, the neighbors are other search users that may influence or be influenced by the central user; however, for an advertiser, neighbors are her competitors who might bid on the same keyword or compete for the same market segment.

The middle part is the control center with direct access to the search-advertising environment, search users' behavior models, advertisers' strategies and the artificial societies of search users and advertisers (i.e., the red dash line). Through the experiment control center, we can design and control computational experiments with different settings of search engine advertising processes and strategies.

The execution of advertisers' strategies depends on processes for bidding auctions in the search engine advertising environment (i.e., the broad arrow). These processes specify how advertisers will be ranked according to their strategies and charged based on search users' activities. Such a simulation system could exhibit dynamic aggregate patterns emerging from large volumes of "local" individual activities by these agents.

### 3.2 The Search User Model

In a search engine advertising session, search users' states are mainly specified by their search interest (i.e., topics to search) and search intent (i.e., actions followed by queries):

- Search Interest: In general, users' search interests are a random variable that is not easy to model. In this study, we notice the correlations between the user's demographic information and their search preferences (e.g., interests) (Purcell 2011; Purcell, Brenner & Raine 2012). Thus, we employ demographic factors (e.g., gender, age, income and education level) to determine types of agents and their behaviors. Such demographic factors constitute the basic innate part of search users' background and are widely used in market segmentation. We employ description logics to describe user information and knowledge at the semantic level, due to their effectiveness in previous research (Yang, Aufaure & Claramunt 2007). Particularly, such a method can handle inconsistency and incompletion issues caused by missing user information (Baader et al. 2003), which is important for our research context. Consider an example, a search user with high educational level has high scholar interests (thus tends to query and click on contents related to scholarly and high-tech subjects) can be described as the follows with description logics:

    $Search\_user \cap \exists hasEduLevel(high) \Rightarrow Search\_user \cap \exists [\$1]hasInterest(\$2:scholar\ \$3:high)$

    The search_user concept is described in Table I. In the current implementation, we extract the mapping rules from the search user's demographical information to interests from the results of the CNNIC surveys. Specifically, based on the definition of search user types by the CNNIC, we can tell the type of a search user according to her demographical information, then get the conditional probability of search user's interest on different topics, upon user types, based on the released statistical results. The rules derived from CNNIC surveys provide a basic representation of China search market's users.

Table I. The Search User Concept

| Concept description | Search_user⊆Person ⊆⊆Thing |
|---|---|
| Basic information | ∩ (≤1hasCode String) |
|  | ∩ (hasAgeInt) |
|  | ∩ (≤1hasGender Gender) |
|  | ∩ (hasIncome Inter) |
|  | ∩ (hasEduLevelEduDegree) |
|  | ∩ (hasQuery String) |
|  | ∩ ( [$1] hasAction $2: Action $3: URL) |
| Derived information | ∩ ( [$1] hasInterest $2:Topic $3: InterestDegree) |
|  | ∩ (hasIntent Intent) |

- Search Intent: Showing interest by querying or clicking does not mean a search user is ready to take further action, such as making a purchase. In previous research, Jansen et al. (2008) showed that search user's intent consists of three automatically identifiable categories: *informational*, *navigational*, and *transactional*. The informational search users aim to find contents relevant to a particular topic to address his information needs. The navigational search users intend to locate a particular website in mind certainly or uncertainty. The transactional search users want to locate a website with specific goals in order to obtain either some items or services of interest. Considering that in our framework, there could be users who do not yet intend to purchase, we add a "latent" state to represent the idle search users. We propose a self-loop state model to describe the change of search users' intents, i.e., latent→ informational→ navigational→ transactional→ latent→... by assuming that a search user will return to the latent state after her transactional requirements are satisfied. Figure 2 visualizes the transition diagram of the four categories of search intents. We expand prior work here by asserting that there is a relationship among these search intent categories. Generally speaking, search users whose search intents are at informational, navigational, and transactional categories have an ascending tendency to take action on the landing page (e.g., register and purchase items, as expected by advertisers). Search users may also take informational queries prior to navigational, for example. In the meanwhile, they also have a probability of jumping back to the latent state. In the current work, we take search intent as an independent factor instead of being determined by states of demographical factors. That is, it evolves in a cycle on its own.

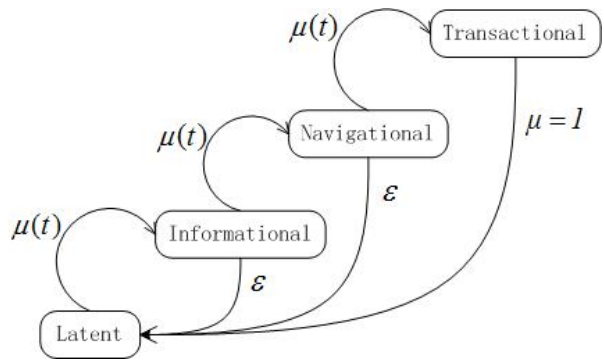

Figure 2. The intent state transition diagram for Search users

Based on the above specifications, the state of a search user can be represented as a four-tuple:

$$S_u = \langle \Gamma, X, \Lambda, \mu \rangle$$

where:

- $\Gamma \in \{sus_p | 1 \leq p \leq P\}$ is the search user type specified by their demographic information. $P$ is the number of search user types defined in a certain application;
- $X = \{topic, level\}$ denotes a search user's interest and level on a topic, which can be inferred from the search user type. The interest level can be either continuous (e.g., $level \in [0, 1]$) or discrete (e.g., $level = \{low, middle, high\}$), $topic \in \{sus_q | 1 \leq q \leq Q\}$. $Q$ is the number of search topics defined in a certain application;
- $\Lambda \in \{sus_s | 1 \leq s \leq S\}$ is the intent of the search user; $S$ is the number of search intent defined in a certain application. In our research, $\Lambda \in \{latent, informational, navigational, transactional\}$.
- $\mu \in [0, 1]$ is the transition probability of states of search intent.

For the values, we leverage Jansen, Booth & Spink (2008) for the classification of search users and survey data for specific search user demographics. For the interest level, we took a simplified probabilistic approach for this research. In the future work, we will extrapolate based on the factors, such as the number of queries submitted or other kinds of user behaviors.

Given the defined search user states, we define the cellular automata of search users (CASU) in the artificial society framework. Without loss of generality, we adopt the two-dimensional cellular automata in this research, where each cell of the automata contains a search user agent. The CASU is given as:

$$CASU = \langle S_U, Z_U, N_U, \varphi_U, f_U \rangle,$$

where, $S_U$ is the finite set of states of search users; $Z_U = \{(i,j) | 1 \leq i \leq I_U, 1 \leq j \leq J_U\}$ is a two dimensional $I_U \times J_U$ lattice of the two-dimensional cellular automata; $N_U$ is the subset of search users in the neighborhood, surrounding a central cell; $\varphi_U : S_U \times S_U^N \to S_U$ is the local state transition function; and $f_U: Z_U \to S_U$ is the initialization function.

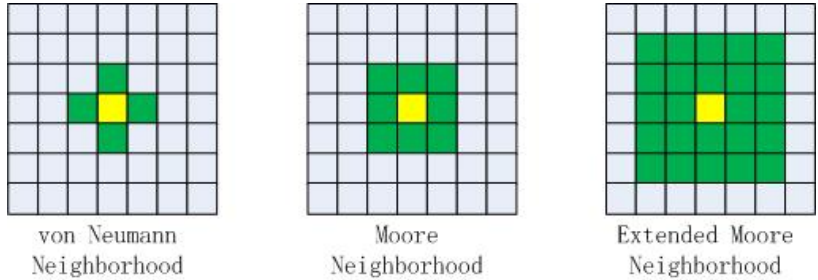

Figure 3. The three cellular automata neighborhoods

In this definition, upon the cellular automata of $Z_u$, $N_u$ defines the scope of search users' interactions using the "neighborhood" concept. For search users, the neighborhood captures the effect of word-of-mouth (WOM) among users in a social network's (Zarrinkalam, Kahani & Bagheri 2018) local community (Goldenberg, Libai & Muller 2001; Jansen et al. 2009; Zwass 2010; Zhang et al. 2017; Hung 2017), where they have influences on search behaviors of each other, thus probably submit similar queries and take similar actions. On a two-dimensional space, the neighborhood can be defined by either the von Neumann space (with 4 neighbors), Moore space (with 8 neighbors), or extended Moore space (with 24 neighbors) surrounding a central cell (as shown in Figure 3).

The interactions among search users are modeled so that each search user's state at the t+1 stage is a function of the states of herself and her neighbors at the t stage, i.e., the state transition function $\varphi_U$ is in the form of:

$$S_{U,t+1}(\alpha) = \varphi_U(S_{U,t}(\alpha), S_{U,t}(N_U(\alpha))),$$

where $\alpha$ denotes the search user in the central cell of CASU, and $N_U(\alpha)$ denotes the set of search users in the neighborhood surrounding $\alpha$. The transition of search users' states is reflected on the search intent[2]. Specifically, at the t stage, the transition probability for an individual search user is automatically increased by an incremental value (e.g., ρ). Besides, a cross-effect value is computed through comparisons with other individuals in the neighborhood. When a neighbor's search intent level is lower than or equal to the focal search user, the cross-effect is zero. When a neighbor's search intent is higher than the focal search user, the cross-effect is specified as three discrete values $\delta_1$, $\delta_2$, and $\delta_3$, based on the difference in terms of search intent, i.e., $\delta(\alpha, \beta) = \delta_{s(\alpha)-s(\beta)}$, where $\beta$ is one of neighbors of $\alpha$. The total cross-effect value is a sum of cross-effects from all neighbors of the focal search user. The overall transition function with respect to search intent is instantiated as a discrete-time-state Markov chain, given as follows:

$$\Pr(\Lambda_{t+1} = s \mid \Lambda_t = s_t) = \mu_{t+1}$$

$$\mu_{t+1} = \mu_t + \rho + \sum_{\beta \in N_U(\alpha)} \delta(\alpha, \beta)$$

By labeling the state space, the transition matrix is:

$$\Pr = \begin{bmatrix} 1 & \mu(t) & \mu^2(t) & \mu^3(t) \\ \varepsilon & 1 & \mu(t) & \mu^2(t) \\ 0 & \varepsilon & 1 & \mu(t) \\ 1 & 0 & 0 & 1 \end{bmatrix}.$$

If the accumulated value of transition probability is larger than 1.0, the transition probability is updated as 1.0. By following the transition matrix, the search user's search intent transitions are simulated. If a search user successfully transits to the next level of search intent, the transition probability is reset as its initial value.

Basically, the initialization function $f_U$ affects the model at t=0, where each individual search user is assigned a random user type and a random search intent following a certain probability distribution (e.g., a discrete uniform distribution) and a transition probability $\mu$. The initialization function ($f_U$) is:

$$f_U = \begin{cases} p(Y = \sup_p) = \frac{1}{p}, \sup_p \sim U(\Gamma) \\ p(Y = \sup_s) = \frac{1}{s}, \sup_s \sim U(\Lambda) \\ \mu \sim U(0, 1) \end{cases}.$$

---

[2] Note that we do not consider the evolution of search user types in this research. That is, the search interest is kept fixed during the CASU evolution. It will be an interesting research issue in the future work.

### 3.3 The Search Advertiser Model

In search engine advertising, an advertiser has one or more products to sell, a budget to expend, a website, several campaigns with bid values for keywords, and one or more strategies to fulfill her advertising goals. Typically, an advertiser has to select a set of keyword phrases relevant to their target market and determines bids on those keywords. Besides, she needs to design advertising campaigns (including adcopy, keywords, and bids on these keywords) following some rules defined by major search engines. The Search_advertiser concept is defined in Table II using description logics in a similar way as the Search_user concept.

Table II. The Search Advertiser Concept

| Concept description | Advertiser ⊆ Person ⊆ Thing |
|---|---|
| Basic information | ∩ (≤1hasCode String) |
| | ∩ (hasBudget Float) |
| | ∩ (hasCampaign Campaign)) |
| | ∩ (hasBid Float) |
| | ∩ (hasProduct Product) |
| | ∩ (hasWebsite URL) |
| | ∩ (hasStrategy Strategy) |
| | ∩ (hasPerformance Performance) |

Furthermore, the cellular automata of search advertisers (CASA) can also be defined in a similar fashion. It is given as:

$$CASU = \langle S_A, Z_A, N_A, \varphi_A, f_A \rangle,$$

where $S_A$ is the finite set of states of advertisers; $Z_A = \{(i,j) | 1 \leq i \leq I_A, 1 \leq j \leq J_A\}$ is a two dimensional $I_A \times J_A$ lattice; $N_A$ is the subset of search advertisers in the neighborhood that can be defined using the von Neumann, Moore, or extended Moore neighborhood surrounding a central cell; $\varphi_A : S_A \times S_A^N \rightarrow S_A$ is the local state transition function; and $f_A : Z_A \rightarrow S_A$ is the initialization function. For advertisers, the neighborhood represents advertisers with interest on a similar set of keywords. A query will trigger the bidding auction process of advertisers within a neighborhood. The entire advertiser space (the grid) represents different types of advertisers focusing on different products/keywords. For the search advertiser's specification, our framework can support complicated rules[3]. The state of a search advertiser consists of her remaining budget and bid price. The former decreases a certain amount determined by the bid price set by the advertiser (i.e., the latter) and the pricing model implemented by the search engine once a search user clicks. Basically, search users are proactive and self-evolving; however, the advertisers are passive and their state transitions are triggered by the queries generated from the search users, through a series of search engine advertising processes defined by the underlying environment.

An advertiser usually aims to maximize different forms of advertising performance. Our framework supports different advertising strategies and generates various statistics to track the performance of advertising strategies;

---

[3] Since we do not have information on advertiser statistics, the experiments are conducted on assumed distributions of advertisers in the current implementation.

however, this research's objective is not to determine an optimal strategy. There are two common criteria used to measure the advertising performance in search engine advertising (e.g., the number of clicks and the cost-per-click because these indexes are directly affected by bidding strategies). The former measures the total website traffic obtained from search engine advertising activities, and the latter indicates the advertising return-on-investment (ROI). In this work, we take a measure that balances a tradeoff between the number of clicks and the cost-per-click (e.g., the advertiser's profit). Particularly, the utility of an advertiser is defined as the total profit expected from advertising activities in search engine advertising. Let $d$ denote the total number of query demands (relevant to an advertiser's promotion activities) in a search market and $c$ the (average) click-through-rate (CTR). Thus, $d \times c$ represents the number of potential clicks that can be obtained by the advertiser. Let $r$ the (average) conversion rate, $v_s$ the advertiser's profit-per-sale, $p$ the (average) cost-per-click, so $r \times v_s - p$ represents the profit generated from a click. Thus, the total profit for an advertiser can be represented as:

$$\text{Profit}_A = d \times c \times (r \times v_s - p).$$

In this research, we do not have a concrete measure of profit per sale. Thus, we let $v_c = r \times v_s$ be the advertiser's profit-per-click and rewrite the utility function for an advertiser as:

$$\text{Profit}_A = d \times c \times (v_c - p).$$

### 3.4 The Search Engine Advertising Environment

The search engine advertising environment defines auction mechanisms, ad retrieval, ranking, and pricing models, and the underlying processes. In the following, we provide major implementation details of search engine advertising processes.

*Ad Retrieval:* When a query is located in a cell of the CASA, we take advertisers with the same topic, relevant to her products or services to promote within the neighborhood boundary as the candidate advertisers or advertisements for the bidding process. The current implementation only considers exact matching of query topics, of search users, and of products or services of advertisers. We reserve other matching options for future research.

*Ranking:* The candidate advertisements are ranked based on the product of its bid times the relevance to the query (i.e., *bid price\*relevance)*. In this research, we use relevance as the proxy for quality score. In effect, relevance is the kernel component to construct the quality score, according to Google Adwords[4]. In the CASA, relevance is computed as proximity, which is the reciprocal of the distance between the advertiser's and the query's position. In the ranked list, these ads differ in prominence effects (Jeziorski & Moorthy 2017).

*Pricing:* The search advertisers pay the search engines according to some specified pricing mechanisms when a search user clicks on her advertisements. It is the so-called pay-per-click (PPC) model. The price of a click is determined by generalized first price (GFP) or generalized second price (GSP). That is, when an advertiser gets a click, as for the former, she pays a price equal to her bid submitted; as for the latter, she pays a price equal to the second highest bid amount in the ranking list. When the budget is used up, the advertiser will leave the auction, and her advertisements will not be shown anymore.

---

[4] https://support.google.com/adwords/answer/140351?hl=en

### 3.5 The Lifecycle of a Search Engine Advertising Session

A search engine advertising session is driven by search users' query, click (on advertisements), and (purchase) actions, along with the advertisers' responses through the underlying processes in the advertising environment.

*Query:* At the individual level, a search user's interest drives query generation. That is, search interest determines which types of products or services might be of interest and to which degree. Once a query is submitted, it is located in a cell of the CASA with equal probability (i.e., 1/q), triggering an auction over a set of relevant advertisers (specified through the ad retrieval process) to rank them and determine the top-n (through the ranking process).

*Click:* The search user then has a probability to click on the listed advertisements. The click-through-rate (CTR) of an ad refers to the probability that it receives a click, given it is included in the sponsored list. Studies on the correlation between ranking positions and click-through-rate indicate that the relationship between click-through-rate and the ranking position is a power law curve, e.g., that the CTR dramatically rises from position 5 to position 1 (Divecha 2012; Jansen, Liu & Simon 2013; SEO-Chat. 2012). The basic cascade click model assumes that users will visually scan the list of ads from the top to the bottom. The idea behind is that a search user begins scanning the list of ads with an initial probability to click, and the click probability degrades at a certain rate during the scanning process (Kempe & Mahdian 2008).

*Action:* On the landing page or some Web page with a few links from it, the advertiser specifies the actions expecting search users to take, e.g., purchase a product, register as a community member, etc. Search users have a probability to take the further action (as expected by the advertiser) based on the judgment if products for sale or services provided satisfy their query tasks. Otherwise, they will quit the page. In both cases, a search engine advertising session is closed.

### 3.6 Framework Instantiation

We implement an Experimental Platform for Search Engine Advertising (EXP-SEA) based on the simulation framework proposed above. We apply a complex system modeling technique called stochastic cellular automata to generate an artificial society of search users and advertisers. In the following, we investigate implementation issues with respect to artificial societies of search users and advertisers.

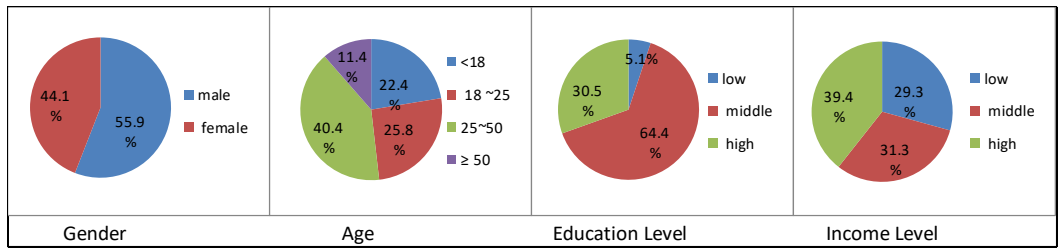

Figure 4. Demographic statistics of Chinese search users (adapted from (CNNIC 2012))

*Search Users:* We build a finite number of artificial search users in a stochastic cellular automata framework by following the characteristics of Chinese search users as reported in the CNNIC surveys (CNNIC 2011b; CNNIC 2011c;

CNNIC 2011a)[5]. For user profiles, we leveraged the CNNIC results to provide specific values. While, for the other parameters of the framework, we could rely on generally supported values in the field.

Specifically, we assign the demographical information of search users (as shown in Figure 4) on gender, age, income, and education level based on statistics reported in the CNNIC survey. We map search users to user types defined by the CNNIC using the demographical information, on which the search preferences are reported in Table III[6]. Table III also reports the probability that users may submit a query on different topics, which is used to generate queries in the EXP-SEA. For example, a user between ages 12 and 18 years with a low-income level and a middle educational level would be considered as a teenager, and 51.6% of teenage search users search for videos.

Table III. User type and the percentage of users submitting queries of a topic

| Search user type | Age | Education | Income | Video | Music | Scholar | Literature* |
|---|---|---|---|---|---|---|---|
| Teenager | 12 ~ 18 | Middle | Low | 51.6% | 50.0% | 30.7% | 35.4% |
| University student | 18 ~ 25 | High | Low | 55.2% | 55.2% | 44.8% | 35.2% |
| White collar | 25 ~ 60 | High | High | 43.2% | 39.9% | 32.3% | 21.0% |
| Blue collar | 18 ~ 60 | Middle | Middle~High | 43.8% | 38.2% | 21.3% | 20.3% |
| Farmer | 18 ~ 60 | Low | Low | 43.2% | 41.3% | 32.2% | 24.5% |

(* Literature means novels, fiction, etc. published online.)

*Search Advertisers:* We build a set of search advertisers that are uniformly distributed over the four topics in Table III. As shown in previous studies, advertisers of major search engines (e.g., Google) show a phenomenon of the long tail, e.g., small advertisers contribute at least half of the revenue for Google (Anderson 2006). Among Chinese advertisers (CNNIC 2011a), 78% of them with annual turnovers ranging from one million to 50 million Chinese Dollars (RMB), which also indicates a power law distribution. The simplest and most common way to make decisions on advertising budgets is to invest a constant percentage of desired sales or profits (Sissors & Baron 2002). In this sense, we reasonably assume that search advertisers' budgets follow a power law distribution. Similarly, we also assume that search advertisers' value-per-click (VPC), which directly determines their bid price, follows a power law distribution.

## 4. Experimental Studies

The EXP-SEA platform could support the design, execution, and control of various computational experiments on search auction mechanisms and advertising strategies in user-defined scenarios and realize the quantitative evaluation and analysis. In this research, we conduct three experiments to showcase the value of our proposed framework. In the first experiment, we evaluate the effect of the eWOM on the search engine advertising performance. In the second experiment, we inspect the impact of advertising competitions on the search market. In the third experiment, we examine the impact of strategic bidding behaviors on the search market.

---

[5] Without loss of generality, our framework can be easily instantiated with flexible user characteristics derived from different sources.
[6] Note that in the current implementation, we do not take into account gender because there is no explicit statistical information about the relationship between the gender and search interests in CNNIC reports; however, our simulation framework is extendable to include more demographic factors and interaction rules with minor adaptations.

### 4.1 General Setup

For experimental purposes, we generate an artificial society including 100 search users and 25 advertisers related to each search topic (i.e., totally 100 advertisers), as specified in Table III[7]. The artificial search users are used as the base to generate queries. This environment can simulate the market with a proportional number of participants. The CASU starts with an initial state of searching intentions, which are randomly assigned to the four states. The demographic characteristics of search users are aligned with Figure 4, while Figure 5 illustrates the distribution of search users over age. The CASA starts with an initial state of budgets and bids following the specification in section 3.3, which then evolves throughout a series of search auctions by taking these queries as inputs. Figure 6 illustrates the distributions of advertisers in terms of budget, which are randomly assigned to the four search topics.

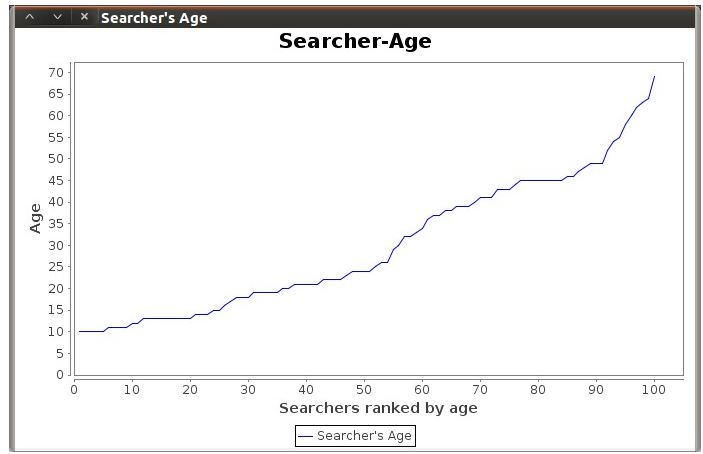

Figure 5. The distribution of search users over age

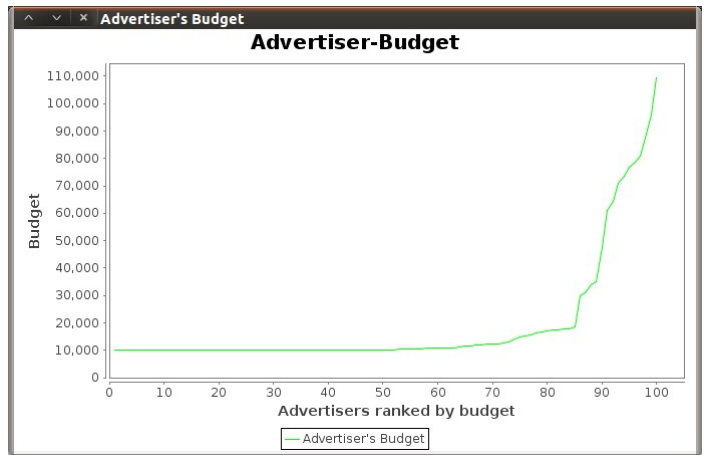

Figure 6. The distribution of advertisers over budget

---

[7] Note that the ratio of search users and advertisers is 4:1 because every search user has certain interests on these four topics. Although this ratio appears to be somewhat less than that in practical situations, it won't lead to a significant difference in terms of advertising performance because we can fully control the number of queries generated from these artificial search users. Basically, search queries are the key input to trigger the CASA and the consequent processes of search engine advertising. Moreover, the difference is not so much if we only consider those who are willing to browse the sponsored list and probably click one or more of them on SERPs. Additionally, since this research mainly focuses on the relative revenue of advertisers, the number of search users does not make a difference in our findings.

During the experiments, each search user will generate a certain number of queries based on their user types. Each query triggers a search engine advertising auction involving several advertisers. If one of these advertisers obtains a click, she is charged a certain price. Due to the lack of behavioral research and empirical data, it is difficult to specify the parameters of a search user's internal status, such as the intent of transition and cross-effect. In the experiments, we simulate 1,000 stages of CASU evolution with the initial transition probability μ = 0.10, the incremental value of transition probability ρ = 0.10, and the cross-effect parameters $\delta_1$ = 0.125, $\delta_2$ = 0.25, $\delta_3$ = 0.45.

Note that our experiments are conducted upon assumed distributions over advertisers; however, extra information from surveys is needed to fully model advertisers' preferences on the market. In the scope of the current research, such information is not available. Since our framework can support detailed specification of advertising preferences, more complicated experiments can be easily conducted with richer advertiser information.

The first experimental study aims to explore the effect of transition parameters (e.g., the eWOM) on the advertising performance. For the other two studies, we assume a reasonable set of transition parameters, and these parameters are kept fixed. In other words, we can explore the effect of competition levels and strategic bidding behaviors given a set of transition parameters (of search users) fixed[8].

### 4.2 Evaluation Metrics

Through the statistics module, our platform can report the following measures of the search market:
- The number of impressions, i.e., the number of times that the advertisement is viewed by the search users.
- The number of clicks, i.e., the total website traffic obtained from search engine advertising.
- The number of actions, i.e., the total actions made by the search users
- The click-through-rate (CTR), i.e., the number of clicks divided by the number of impressions.
- The conversion rate (CR), i.e., the number of actions divided by the number of clicks.

Furthermore, we also inspect metrics related to the search providers' and advertisers' economic benefits. Through the system, we can measure an advertiser's profit as described in Section 3.3. In this research, we assume that each click can raise the same amount of profit (i.e., $v_c$) for all advertisers, and its lower bound is the largest bid price by advertisers within a topic so that the net profit per click will be positive for all advertisers. Thus, we take this lower bound as the estimation of $v_c$ to calculate advertisers' profits, i.e., $v_c = b_{max}$.

These evaluation metrics are implemented in the EXP-SEA as part of the simulation platform and can be used to assess the performance of advertising strategies. In this study, we report the average values of these metrics to measure the search market's performance.

### 4.3 Experimental Study 1: WOM's Impact on the Search Market

The first experiment aims to show the framework's ability to examine the effect of the WOM on the search market. In our framework, the eWOM is reflected by the cross-over effect within the neighborhood of search users in the CASU. We compare the results without and with the cross-over effects. In this experiment, all advertisers take a static bidding strategy, i.e., their bidding prices are kept unchanged during the evolution period of the CASU. In the setting with the eWOM effect, we assign values for the cross-effect parameters (i.e., $\delta_1 = 0.125$, $\delta_2 = 0.25$, $\delta_3 = 0.45$); while in the setting without considering the eWOM effect, we assign a zero-value for the cross-effect parameters (i.e., $\delta_1 = 0.0$, $\delta_2 = 0.0$, $\delta_3 = 0.0$). The results of the settings with and without eWOM are shown in Table IV.

---
[8] We believe that a set of fixed parameters for search users forms an experimental soil for evaluating search advertisers will not confuse the comparison studies in these two case studies.

Table IV. Advertising performance with and without eWOM

|  | With eWOM | Without eWOM |
|---|---|---|
| Impression | 239,503 | 215,793 |
| Click | 11,653 | 10,207 |
| Action | 3,977 | 2,294 |
| Click-through-rate | 0.050 | 0.048 |
| Conversion rate | 0.351 | 0.225 |
| Cost-per-click | 11.623 | 11.700 |
| Market profit | 367,207 | 308,617 |

From Table IV, we notice that all advertising indexes (i.e., number of impressions, clicks, and actions, the click-through-rate, the conversion rate, and the cost-per-click) are larger in the setting with the eWOM effect than without the eWOM. As we mentioned in Section 3.2, the eWOM effect only works on the evolution of search intents that determine the probability of purchase actions in a search engine advertising session. Thus, we can easily explain the difference in terms of the number of actions and the conversion rate between the settings with and without the eWOM effect. While, the difference of other advertising indexes is attributed to the aggregate effect of random factors in the search engine advertising processes, such as the query generation and click model. Moreover, we also see that the market profit is bigger while considering the eWOM effect. It implies that the eWOM has a significantly positive effect on the market profit in search engine advertising[9].

### 4.4 Experimental Study 2: Competition Intensity's Impact on the Search Market

The second experiment aims to show the framework's ability to examine the competition level's effect on the search market. In our framework, the competition level among advertisers is reflected by the number of advertisers involved in each auction process (i.e., the size of the neighborhood space in the CASA) triggered by a given query generated from the CASU. We experiment with three types of neighborhood spaces: the von Neumann space (with 4 advertisers in each auction), the Moore space (with 8 advertisers), and the extended Moore space (with 24 advertisers). Similarly, in this experiment, all advertisers take a static bidding strategy, i.e., their bidding prices are kept unchanged during the evolution period of the CASU. The results of settings with three competition levels are shown in Table V.

Table V shows that, a higher competition level leads to a larger number of impressions, clicks, and actions. One major reason for this phenomenon is that the lower competition levels might not fully cover the slots provided by search engines. Although the CTR decreases at the higher competition level, the CR shows a tendency to increase with more intensive competitions. In other words, if a user clicks an item from a long list, there is a higher chance she will eventually make the purchase, as compared to a click from a short list. The CPC is reduced at the higher competition level, implying that an advertiser needs to pay less to be included in the sponsored list. However, it's more difficult for her to get traffics from search engine advertising campaigns at the higher competition level.

---

[9] Note that the utility function of search advertisers (given in Section 3.2) is somewhat conservative in that the value-per-click (VPC) is kept fixed. In practice, for an advertiser, with the conversion rate and the number of actions increasing, she can get a bigger VPC.

Table V. Advertising performance at three competition levels

|  | Competition level-1 (von Neumann) | Competition level-2 (Moore) | Competition level-3 (Extended Moore) |
|---|---|---|---|
| Impression | 91,464 | 156,803 | 239,503 |
| Click | 5,824 | 8,770 | 11,653 |
| Action | 1,939 | 3,005 | 3,977 |
| Click-through-rate | 0.065 | 0.056 | 0.050 |
| Conversion rate | 0.334 | 0.341 | 0.351 |
| Cost-per-click | 14.902 | 12.120 | 11.623 |
| Market profit | 1,59,715 | 250,034 | 367,207 |

Table V reports the market performances with respect to three competition levels. From the experiments, the market profit increases from 159,715 to 250,034, then to 367,207 with the increase of competition levels. This implies that a higher competition level can increase the market-level profit. According to the market efficiency literature (Färe et al. 2004), the competition increases the market efficiency, which eventually improves advertising effectiveness. In this sense, as market makers, search engines have an incentive to increase the advertising competition on every keyword.

The findings of our experiment fit the theoretical projections on a perfectly competitive market (PCM) in microeconomics. The search engine advertising market nicely satisfies two essential assumptions of a PCM (Mankiw 2012) where 1) the goods for sale are all almost the same (i.e. clicks on a set of keywords), and 2) no single buyer (i.e., advertiser who buy the clicks) or seller can dominate the market price due to the numerous search users, advertisers, and keywords. In a PCM, the more competition leads to the more Pareto efficient resource allocation, thus resulting in a greater number of goods and lower prices (Brownstein 1980; Stiglitz 1981). Considering the competition in the context of search engine advertising, the increasing number of advertisers implies that the demand (for clicks) is increased. If the supply (i.e., search users' clicks) remains the same, it leads to a higher equilibrium price; however, in our experiments, the number of search clicks also increases, which may be due to more keywords (or more sponsored links for a keyword) made available by the increased number of advertisers. As a result, more clicks on (more) keywords of interest are delivered in search engine advertising, and the overall price is lowered.

With the help of our simulation platform, search engines can conduct similar simulations based on the characteristics of the search market. Search engines can then control the number of valid advertisers (or attract more advertisers) to maintain a healthy market competition, which will not only benefit the search engine but will also benefit small advertisers by providing a lower cost (Hunt & Morgan 1995).

### 4.5 Experimental Study 3: Advertising Strategies' Impact on the Search Market

In the third experiment, we intended to show our frameworks' ability to examine strategic bidding behaviors' impacts on the search market. In this study, we mix two types of strategies in the experiments. The first strategy is a static bidding strategy where the bid prices are fixed during the CASA evolution. This is a common method in practical search engine advertising situations because advertisers usually do not have sufficient knowledge and time for real-time advertising operations. The second strategy is a dynamic, heuristic-rule-based bidding strategy taking the complete-information assumption that an advertiser is aware of other advertisers' bid prices. At each step, there is a list of bid prices set by advertisers in a specific bidding game. After the t-th step, each advertiser is entitled to adjust her bid price according to the observed information of bid prices. That is, the advertiser will adjust her bid price based on how much her bid price deviates from the mainstream (i.e., the average bidding prices). If the absolute value of her personal deviation is larger than the standard deviation of the list of bid prices, she will adjust her bid prices toward the average value of the list. The deviation degree determines the adjusting magnitude. In essence, this strategy captures a herding effect among advertisers. The details of this dynamic bidding strategy are given in Algorithm 1.

---

**Algorithm 1** (The Dynamic Bidding Strategy)

Input: $b_{i,t}$ (the bid price by advertiser $i$ at time $t$), **b** (the bidding vector of $n$ advertisers at time $t$), $\kappa$ (the adjustment scale).

Output: $b_{i,t+1}$ (the adjusted bid price by advertiser $i$ at time $t + 1$).

1) Compute the standard deviation of **b**: $\sigma$, and the average value of **b**: $\bar{b}$;
2) Compute the personal deviation for advertiser $i$: $\xi_i$;
3) If $(\xi_i - \sigma) \leq 0$, then $b_{i,t+1} = b_{i,t}$; otherwise:
    a) if $(b_{i,t} - \bar{b}) > 0$, then $b_{i,t} = b_{i,t} - |b_{i,t} - \bar{b}|/\kappa$;
    b) if $(b_{i,t} - \bar{b}) < 0$, then $b_{i,t} = b_{i,t} + |b_{i,t} - \bar{b}|/\kappa$;

---

By changing the ratio of advertisers employing the two strategies, we create 5 scenarios with 0%, 25%, 50%, 75%, and 100% strategic bidding behaviors (i.e., taking the dynamic bidding strategy), respectively. The simulation results are illustrated in Figure 7.

Figure 7 illustrates the market-level performance in terms of impressions, clicks, and profits with respect to the five scenarios under four topics (i.e., literature, music, scholar, and video). As shown in Figure 7, when strategic bidding behaviors in a search market are less than 50%, the market profit generally remains stable with a small fluctuation. When strategic bidding behaviors in a search market are more than 75%, the market profit continues dropping.

The bidding strategy described in Algorithm 1 has two main impacts on advertisers in search engine advertising. First, an advertiser with a much higher bid price can always win a higher position in each auction but with a higher CPC. This strategy can help her reduce cost-per-click to a reasonable level by gradually reducing her bid price. Second, for an advertiser with a much lower bid price who usually loses in most auctions (or only get a much lower position), employing this strategy leads to more clicks for her by gradually increasing her bid price to a reasonable level.

Since the bid price in our experiments follows a power law distribution, there are more advertisers with low bid prices than advertisers with high bid prices. When there are less strategic bidding behaviors (e.g., 25%) in a search market, the bid prices by advertisers will converge to a range specified by advertisers who take a static strategy;

however, if the proportion of strategic bidding behaviors increases significantly, most advertisers have to change their prices; the small number of advertisers with high initial prices will quickly reduce their prices before a large number of advertisers with low initial prices increase their prices. As a result, the overall market price will converge at a lower level and leads to an overall low market profit.

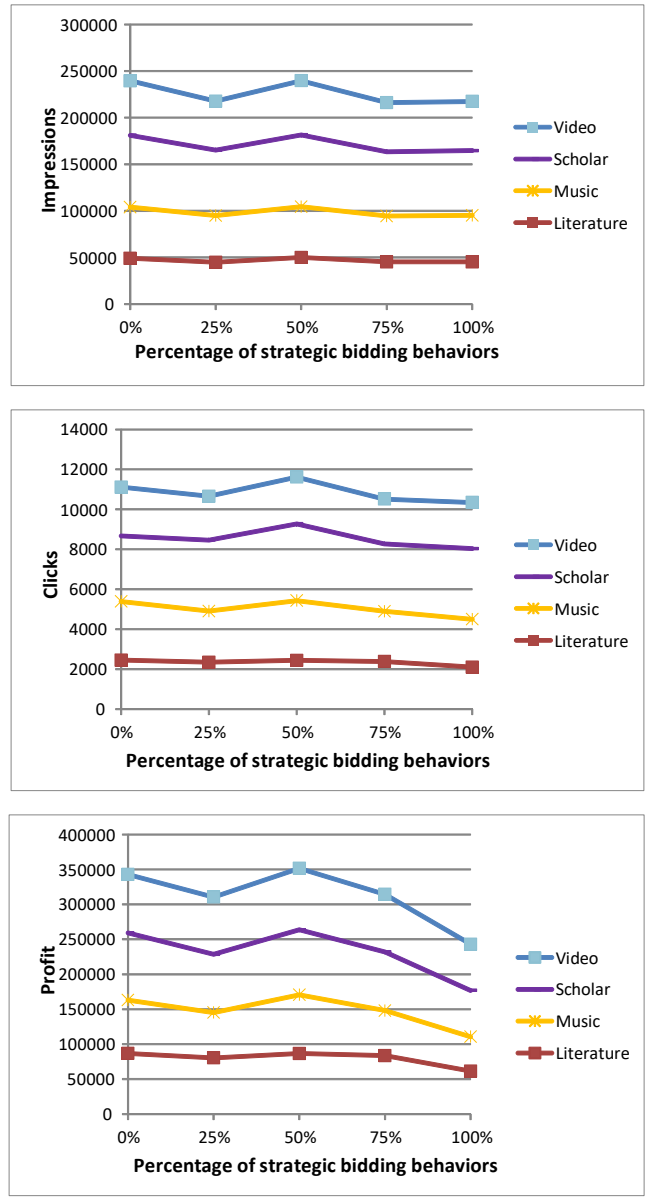

Figure 7. Advertising performance of strategic bidding behaviors

The experimental results indicate that an advertiser should carefully assess her bidding strategy before implementing it, and the assessment should consider the probability of other advertisers taking various strategies. With the help of our simulation platform, search engines can also conduct similar simulations based on characteristics of search markets, and then provide tools or policies on certain bidding strategies to maintain a more profitable market.

## 5. Conclusions and Future Work

This paper presents an artificial society-based simulation framework to support experimental studies of group advertising behaviors by capturing the interactions of search users and advertisers in search engine advertising. We implement an Experimental Platform for Search Engine Advertising (EXP-SEA) and conduct three experimental studies. Experimental studies on the WOM, advertising competitions, and strategic bidding strategies generate interesting results and provide insights into the management of search engine advertising campaigns and the underlying environments. In addition, experimental results nicely fit existing findings and theories, which shows the validity of our framework. Our framework entitles search engines and advertisers to conduct computational experiments to evaluate possible effects of search auction mechanisms and advertising strategies.

The proposed simulation framework has significant methodological and practical implications. At the methodological level, our framework captures interactions among search users and advertisers while controlling the simulation process at a reasonable complexity. For example, it allows us to test the validity of models considering an advertiser's game and search users' eWOM. In the existing literature, such factors are mainly captured using analytical models, which often pose some rigid assumptions and limit the applicability of their conclusions.

At the practical level, our simulation framework, which certainly has some room to improve, shows potentials for developing search engine advertising decision support systems. As shown in our second experimental study, search engines can employ this framework to derive altering market competition plans. As shown in our third experimental study, advertisers can employ the framework to predict the outcomes of putting a certain bidding strategy into practice. There is a rapid growth of search engine advertising market, as many small and medium enterprises (SMEs) are increasingly joining this market. Note that SMEs generally do not have the ability to develop comprehensive decision models or advertising strategies; it would be beneficial for search engines to set up such simulation platforms with parameters estimated from real data to support them. Providing such tools may help SMEs explore various advertising strategies in search engine advertising and potentially increase the overall market efficiency.

In future work, we will extend our framework in the following ways. First, we will further improve the modeling of search users (Liu & Jansen 2018) and advertisers to better reflect their internal states and decision processes, and will calibrate different parameters in our simulation framework. Second, we will investigate the market effect of more complicated advertising strategies and will include benching against a gold standard to evaluate the effectiveness of our framework. Third, we will improve the framework to consider the richer search engine advertising structure and possible interactions with other advertising channels (Joo, Wilbur & Zhu 2016; Kireyev, Pauwels & Gupta 2016; Hasanain & Elsayed 2017), which will lead to complicated search engine advertising auction processes and advertising strategies. Our ultimate goal is to build a comprehensive and effective framework that can support search engines and advertisers' decision-making processes in search engine advertising.

**Acknowledgement**

We are thankful to anonymous reviewers who provided valuable suggestions that led to a considerable improvement in the organization and presentation of this manuscript. This work was partially supported by NSFC (National Natural Science Foundation of China) grants (71672067, 71272236, 91646121).